\documentclass[conference,a4paper]{IEEEtran}
\IEEEoverridecommandlockouts

\usepackage[hidelinks]{hyperref}
\usepackage[cmex10]{amsmath}%American Math Society(AMS) math formatting
\usepackage{amssymb,amsfonts}%AMS extra symbols and fonts
\usepackage{dblfloatfix}%fix double column figure ordering and placement

\usepackage[ruled,vlined]{algorithm2e}
\usepackage{graphicx}
\graphicspath{{Figures/PDF/}{Figures/PNG/}}

\usepackage{booktabs}
\usepackage{siunitx}
\usepackage[numbers,compress]{natbib}
\usepackage{texnames}
\usepackage{bm,bbm}
\usepackage{orcidlink}
\usepackage{acronym}
\usepackage{array}

\newcolumntype{C}[1]{>{\centering\arraybackslash}m{#1}}

\begin{document}

\title{\uppercase{Assessing the Added Value of Onboard Earth Observation Processing with the IRIDE HEO Service Segment}
}

\author{%
\IEEEauthorblockN{%
Parampuneet K.~Thind\IEEEauthorrefmark{1}\IEEEauthorrefmark{2}\IEEEauthorrefmark{3},
Charles Mwangi\IEEEauthorrefmark{1}\IEEEauthorrefmark{3},
Giovanni Varetto\IEEEauthorrefmark{3}\IEEEauthorrefmark{6},\\
Lorenzo Sarti\IEEEauthorrefmark{2},
Andrea Papa\IEEEauthorrefmark{7},
Andrea Taramelli\IEEEauthorrefmark{3}\IEEEauthorrefmark{5}}
\IEEEauthorblockA{\IEEEauthorrefmark{1}Sapienza Università di Roma, Rome, Italy}
\IEEEauthorblockA{\IEEEauthorrefmark{2}Argotec Srl, Turin, Italy}
\IEEEauthorblockA{\IEEEauthorrefmark{3}Istituto Universitario di Studi Superiori (IUSS), Pavia, Italy}
\IEEEauthorblockA{\IEEEauthorrefmark{5}Institute for Environmental Protection and Research (ISPRA), Rome, Italy}
\IEEEauthorblockA{\IEEEauthorrefmark{6}Università di Milano-Bicocca, Milan, Italy}
\IEEEauthorblockA{\IEEEauthorrefmark{7}European Space Agency (ESA), ESRIN, Rome, Italy}
\IEEEauthorblockA{%
Email: parampuneet.thind@uniroma1.it,
charles.mwangi@uniroma1.it,
g.varetto@campus.unimib.it,\\
\hspace*{1.2em}lorenzo.sarti@argotecgroup.com,
andrea.papa@esa.int,
andrea.taramelli@iusspavia.it}
}

\maketitle
% acronyms.tex

\acrodef{AI}{Artificial Intelligence}
\acrodef{EO}{Earth Observation}
\acrodef{HEO}{Hawk for Earth Observation}
\acrodef{EFFIS}{European Forest Fire Information System}
\acrodef{CEMS}{Copernicus Emergency Management Service}
\acrodef{CLMS}{Copernicus Land Monitoring Service}
\acrodef{PNRR}{Piano Nazionale di Ripresa e Resilienza}
\acrodef{SVC}{Service Value Chain}
\acrodef{PNC}{Piano Nazionale Complementare}
\acrodef{ESA}{European Space Agency}
\acrodef{ASI}{Agenzia Spaziale Italiana}
\acrodef{LEO}{low Earth orbit}
\acrodef{IRIDE}{International Report for an Innovative Defence of Earth}
\acrodef{FOS}{Flight Operation Systems}
\acrodef{PDGS}{Payload Data Ground Segment}
\acrodef{MMU}{Minimum Mapping Unit}
\acrodef{RDA}{Rapid Damage Assessment}
\acrodef{SVC}{Service Value Chains}
\acrodef{DPC}{Dipartimento della Protezione Civile}
\acrodef{MIPAF}{Italian Ministry of Agriculture and Forestry Policies}
\acrodef{COTS}{Commercial Off-The-Shelf}
\acrodef{VPU}{Vision Processing Unit}
\acrodef{CMPM}{Central Mission Planning Module}
\acrodef{PDP}{Payload Data Processor}
\acrodef{OCDH}[OBC\&DH]{Onboard Computer and Data Handling}
\acrodef{PDP}{Payload Data Processor}
\acrodef{GSD}{Ground Sampling Distance}
\acrodef{NAS}{Neural Architecture Search }

\begin{abstract}
Current operational \ac{EO} services, including those provided through the \ac{CEMS}, the \ac{EFFIS} (within the \ac{CEMS}), and the \ac{CLMS}, rely primarily on remote, ground-based processing pipelines. While these systems deliver mature large-scale information products, they remain constrained by downlink latency, bandwidth availability, and limited capability for autonomous prioritisation of observations. The \ac{IRIDE} programme is a national Earth observation system launched by the Italian government to support public authorities through timely, objective information derived from spaceborne data. Rather than a single satellite constellation, \ac{IRIDE} is structured as a \emph{constellation of constellations}, integrating heterogeneous sensing technologies into a unified service-oriented architecture. Within this framework, \ac{HEO} generates onboard data products that enable information extraction earlier in the processing chain. This paper examines the limitations of ground-only service architectures and evaluates the potential added value of onboard processing at the operational service level. The \ac{IRIDE} burnt-area mapping service is adopted as a representative case study to illustrate how onboard intelligence can be used to target higher spatial detail (sub-three-metre ground sampling distance), smaller detectable event sizes (a minimum mapping unit of three hectares), and improved overall system responsiveness. Rather than replacing existing Copernicus-based services, the \ac{IRIDE} \ac{HEO} capability is positioned as a complementary layer providing image-driven pre-classification to support downstream emergency and land-management workflows. By situating the \ac{IRIDE} \ac{HEO} service segment within the broader evolution of onboard processing technologies, this work demonstrates the operational value of onboard intelligence for emerging low-latency, intelligence-supported Earth observation service architectures.
\end{abstract}


\section{Introduction}

Operational \ac{EO} services are dominated by remote, ground-based processing pipelines in which full-scene imagery is downlinked before any information extraction occurs~\cite{kaku2013sentinel}. Such architectures traditionally assume that high-level data products must be generated through extensive preprocessing chains executed on the ground~\cite{Meoni2025E2E}. Core European services such as \ac{CEMS} (including \ac{EFFIS}), and \ac{CLMS} will take advantage of this \emph{downlink-first} paradigm~\cite{EffisRda,CLMSsite}. While this approach has proven effective for continental-scale monitoring~\cite{BARBOSA2006S218}, it inherently ties product timeliness to downlink availability and ground-segment throughput~\cite{fang2023service}. As a consequence, such architectures are prone to latency bottlenecks and bandwidth saturation during crisis events~\cite{Chien2017}, motivating hybrid onboard--ground processing concepts aligned with next-generation hybrid constellation and NewSpace mission architectures~\cite{chien2004eo,Madry2018,Pastena2019}.

At the same time, the EO domain is experiencing rapid growth in data volume, spatial resolution, and revisit frequency, intensifying pressure on ground infrastructures~\cite{SELVA201250}. In response, concepts derived from terrestrial edge processing where computation is relocated closer to the data source to reduce latency and data movement~\cite{SHI2016} are increasingly extended to space systems as \emph{orbital edge computing}~\cite{8674608}. In this context, onboard machine inference is emerging as a core architectural element for future \ac{EO} missions~\cite{Denby2023}.

Over the last decade, multiple missions have demonstrated the technical feasibility of onboard \ac{AI} for \ac{EO}~\cite{Giuffrida2020}. 
The $\Phi$Sat-1 mission implemented the CloudScout deep neural network for onboard cloud detection~\cite{CloudScout2020}. 
CloudScout operated on hyperspectral imagery acquired by the HyperScout-2 imager~\cite{Esposito2019Hyperscout}. 
Onboard inference was executed on the Intel Movidius Myriad-2 vision processing unit, a low-power \ac{COTS} accelerator used for in-orbit deployment~\cite{intel_movidius_myriad_2}. 
Building on this line of development, $\Phi$Sat-2 extends the concept toward a multi-application onboard-\ac{AI} platform for operationally relevant tasks~\cite{Melega2025}. 
The mission concept has been presented as supporting applications such as cloud detection and image-to-map conversion~\cite{Marin2021}. 
More recent descriptions further include capabilities such as vessel awareness, wildfire detection, anomaly detection, and onboard compression~\cite{2024AGUFMIN31D2039L}.
Additional recent missions, including real-time fire detection on the FOREST-2 satellite, further illustrate the increasing maturity of onboard intelligence for time-critical event monitoring~\cite{ororatech}. 
At the system level, roadmapping efforts on spaceborne data handling identify onboard processing as a key enabling technology for future autonomous missions~\cite{Furano2018}. 
These roadmaps further highlight the role of \ac{AI} in supporting increased onboard autonomy and responsiveness under downlink constraints~\cite{Furano2020}. 
Complementary technology-focused assessments emphasise onboard \ac{AI} as a driver for next-generation spacecraft data systems~\cite{Pastena2019}.

Despite these advances, most in-orbit AI applications remain technology-driven demonstrators focused on algorithm validation and hardware feasibility~\cite{li2018onboard,dai2019semisupervised}. As a result, onboard AI is rarely embedded within institutional EO service chains with formally defined product specifications, accuracy targets, and operational end users~\cite{ruuvzivcka2022ravaen}. The \ac{IRIDE} programme addresses this gap through a user-driven, service-oriented systems engineering approach for a national, multi-platform EO system~\cite{Costa2023}, in which institutional requirements such as target spatial resolution, revisit time, latency, and thematic accuracy are formalised at service level and translated into constellation architectures and sensing strategies through upstream, model-based design and optimisation studies spanning multiple thematic domains~\cite{IRIDESAR2023,UpstreamOptimization2022,rs12081286}.

Within this framework, the \ac{HEO} microsatellite constellation provides agile multispectral optical sensing at sub-three-metre spatial resolution from \ac{LEO}, operating at an altitude of approximately 500--600~km, together with onboard computing resources dimensioned for AI-based in-orbit processing~\cite{HEO2023}. Unlike previous experimental platforms, HEO is conceived as an operational component of the \ac{IRIDE} service portfolio~\cite{FilippiAiello2024IRIDE}. Burnt-area mapping is a representative service targeting national-scale binary products with a three-hectare minimum mapping unit and $>95\%$ thematic accuracy, supporting institutional users and service evolution pathways identified for Copernicus-based monitoring systems~\cite{Schiavon2021JEM,Schiavon2023JGR}.

This paper evaluates the added value of onboard processing at the \emph{operational service level} using the \ac{IRIDE} \ac{HEO} onboard service segment as a representative national-scale case. Specifically, we: \emph{(i)} analyse the structural limitations of remote-only processing in existing operational \ac{EO} services \emph{(ii)} describe the \ac{IRIDE} \ac{HEO} architecture and its role within the institutional \ac{SVC} and \emph{(iii)} assess the operational implications of onboard intelligence through burnt-area mapping service. The objective is to demonstrate, through an end-to-end onboard processing chain, the tangible service-level benefits enabled by onboard \ac{AI} within next-generation national \ac{EO} architectures.

\section{Service-Level Comparison of Operational Burnt-Area Systems}

\subsection{Structural Limitations of Remote-Only Architectures}

All current large-scale operational burnt-area services rely on a \emph{remote-only} processing paradigm in which full-resolution imagery is systematically downlinked prior to analysis, as exemplified by the \ac{CEMS} \ac{EFFIS} service. While effective at continental and multi-national scale, this architecture intrinsically couples product timeliness to downlink availability and ground-processing capacity. Furthermore, medium-resolution sensing baselines inherently constrain the minimum mapping unit, limiting the detection of small and fragmented burn scars that are often operationally relevant at national and local scale.

\subsection{Comparison Between Copernicus Services and the \ac{IRIDE} National Burnt-Area Service}

Table~\ref{tab:service_comparison} compares the service-level characteristics of selected operational European burnt-area mapping systems with the national burnt-area mapping service developed within the \ac{IRIDE} programme and enabled by the \ac{HEO} optical constellation. The comparison focuses explicitly on downstream service characteristics, distinguishing between acquisition strategies and product generation logic, rather than on upstream mission tasking or platform autonomy. In particular, the comparison highlights overall timeliness as a key differentiating factor, showing how the \ac{IRIDE} \ac{HEO} service segment reduces the time to first actionable information through onboard processing and hybrid product generation.

The \ac{EFFIS} service provides systematic, European-scale burnt-area products derived from medium-resolution satellite data, optimised for seasonal reporting and continental monitoring. Product generation follows periodic processing cycles executed entirely on the ground. The \ac{CEMS} Emergency Mapping component, while not a dedicated burnt-area monitoring service, delivers on-demand crisis products activated in response to specific events and relies on heterogeneous data sources, including commercial very-high-resolution imagery. The \ac{CLMS} supports long-term land-cover and land-use monitoring through systematic acquisitions and multi-annual update cycles, rather than near-real-time fire event characterisation.

By contrast, the \ac{IRIDE} burnt-area service is conceived as a national operational service based on systematic satellite acquisitions, in which product generation is dynamically driven by detected fire events. In this context, fire occurrences identified through external monitoring systems and institutional information flows trigger dedicated processing chains on newly acquired imagery, enabling rapid generation of preliminary burnt-area information without relying on changes to nominal acquisition plans or autonomous satellite retasking. The service targets three-metre spatial resolution, a \ac{MMU} of three hectares, and full territorial coverage under institutional operational priorities. Unlike the other services considered, \ac{IRIDE} integrates onboard processing as part of its operational value chain, enabling early information extraction prior to full data downlink.

\begin{table}[t]
\renewcommand{\arraystretch}{0.9}
\setlength{\tabcolsep}{2.2pt}
\footnotesize
\centering
\caption{Service-level comparison of operational burnt-area mapping systems, with emphasis on overall timeliness}
\label{tab:service_comparison}

\resizebox{\columnwidth}{!}{%
\begin{tabular}{p{2.1cm}cccc}
\toprule
\textbf{Figure of merit} & \textbf{EFFIS} & \textbf{CEMS EM} & \textbf{CLMS} & \textbf{\ac{IRIDE} (HEO)} \\
\midrule
Acquisition mode        & Systematic & On-demand & Systematic & Systematic \\
Product triggering     & Periodic   & Crisis    & Periodic   & Event-driven \\
Processing location    & Ground     & Ground    & Ground     & Hybrid \\

Spatial resolution     & 20 m       & 10--30 m$^\ast$ & 10--100 m & $\sim$3 m \\
Minimum mapping unit   & $\sim$10 ha & --        & $\sim$0.5--1 ha & 3 ha \\

\midrule
\textbf{Time to first info}  & Days       & Hours    & Months--years & Minutes--hours (target) \\
\textbf{End-to-end latency} & Days       & Hours--days & Years    & Hours (target) \\
Downlinked data volume & Full scene & Full scene & Full scene & Reduced (ROI / thematic) \\

\midrule
Onboard processing     & No         & No        & No         & Yes \\
Operational scope      & European   & Crisis    & European   & National \\
\bottomrule
\multicolumn{5}{l}{$^\ast$Sub-metric achievable with commercial VHR imagery.}
\end{tabular}
}
\end{table}

\section{The \ac{IRIDE} Constellation and the \ac{HEO} Onboard Service Segment}

\subsection{The \ac{IRIDE} Programme and National Service Architecture}

The \ac{IRIDE} initiative, developed by the \ac{ESA} with the support of \ac{ASI}, represents Italy's flagship national \ac{EO} programme under the \ac{PNRR} \cite{ESA_IRIDE_mission}. Designed as a multi-segment, end-to-end system, \ac{IRIDE} integrates: (i) an \textbf{upstream segment} of six satellite constellations (optical, SAR, hyperspectral), (ii) a \textbf{downstream segment} consisting of ground infrastructure for satellite operations, data processing, mission planning, and (iii) a \textbf{service segment} dedicated to delivering geospatial products and thematic services to institutional and commercial users \cite{IRIDE_brochure}.

\begin{figure}[t]
    \centering
    \includegraphics[width=\columnwidth]{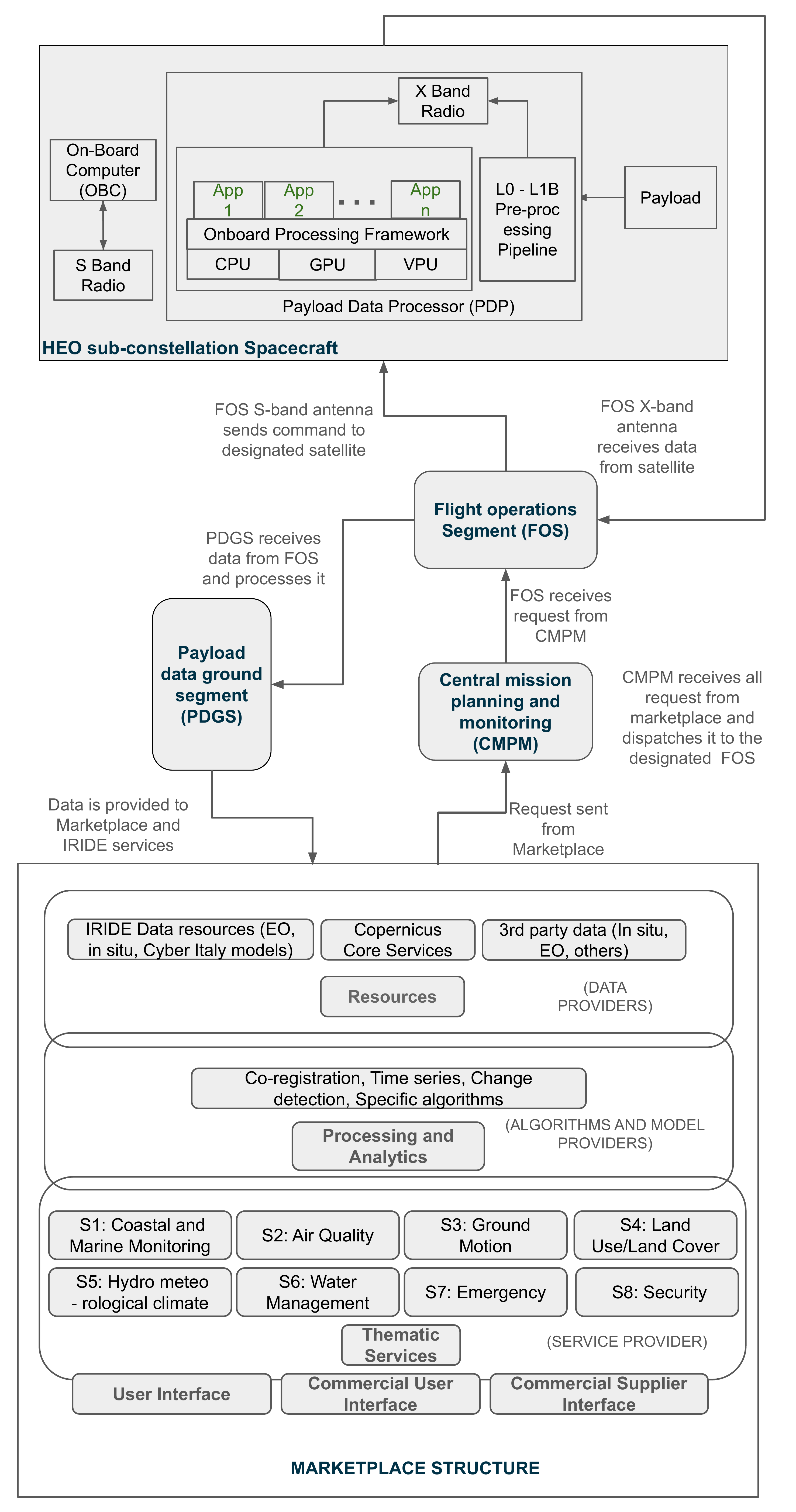} % Replace with your updated diagram
    \caption{System-level architecture of \ac{IRIDE}'s three core segments: upstream (spaceborne sensors including HEO), downstream (FOS, CMPM, PDGS), and service (marketplace-enabled user delivery). Arrows indicate command and data flow paths.}
    \label{fig:IRIDE_segments}
\end{figure}

The upstream segment comprises multiple satellites with diverse sensors. The downstream segment hosts critical systems such as the \ac{FOS}, \ac{PDGS}, and \ac{CMPM}. These components interact to orchestrate satellite tasking, receive insights or raw data, and distribute final products to users via the \ac{IRIDE} marketplace. Figure~\ref{fig:IRIDE_segments} illustrates the \ac{IRIDE} system architecture, with the upstream spacecraft and onboard processing elements reflecting the HEO platform developed by Argotec, the downstream service and marketplace components aligned with the official architecture described in the \ac{IRIDE} programme brochure~\cite{IRIDE_brochure}.

\subsection{The \ac{HEO} Optical Segment: Onboard-Ready Satellites}

Among \ac{IRIDE}'s constellations, the \ac{HEO} optical micro constellation carries high-resolution multispectral sensors (sub-3~m GSD) and has the potential to perform first-order inference directly onboard, enabling real-time responsiveness to events such as wildfires. Each HEO unit is equipped with the Unibap iX5-106 SpaceCloud\textsuperscript{\textregistered} platform, hosting the \ac{PDP} and the \ac{OCDH} subsystems. The \ac{PDP} is responsible for executing AI inference using the Intel\textsuperscript{\textregistered} Myriad X \ac{VPU}, while the OBC\&DH remains responsible for platform control and system monitoring, and the PDP supports it by executing computationally intensive tasks, including onboard image processing. The onboard architecture supports multiple radio interfaces: the S-band for telemetry and telecommand (TM/TC), and the X-band for payload data transmission.

\subsection{Integrated Tasking and Product Delivery Workflow}

The downstream segment includes multiple \ac{FOS} instances, each associated with a specific upstream constellation and responsible for managing satellite health, telemetry, and link control, including payload data downlink. Within this architecture, a dedicated \ac{FOS} is assigned to the \ac{HEO} optical constellation and is operated directly by Argotec. Each \ac{FOS} interfaces with the \ac{CMPM} to receive observation requests and tasking updates, which are then dispatched to the corresponding satellites. Following acquisition, raw data or onboard inference products are downlinked to the corresponding \ac{FOS} and subsequently forwarded to the \ac{PDGS} for ingestion, processing, and distribution through the \ac{IRIDE} marketplace.

This distributed operational setup enables a hybrid service chain in which inference products may be generated onboard the satellite or derived on ground from raw data, depending on data quality, cloud conditions, and service urgency. The architectural integration of constellation-specific \ac{FOS} instances with the \ac{CMPM} and the downstream product management chain ensures end-to-end traceability from tasking to product delivery, as well as timely responsiveness for high-priority applications such as civil protection.

\section{Added Value of Onboard Processing for Burnt-Area Mapping}

Integrating onboard processing in \ac{IRIDE} enables service level gains beyond incremental ground-segment scaling.

\subsection{Latency and Bandwidth Reduction}

By performing radiometric pre-processing and first-order inference directly onboard, the system can downlink thematic outputs or prioritised regions of interest instead of complete raw image strips. This reduces downlinked data volume and alleviates bottlenecks in large-scale fire scenarios. Critically, the end-to-end latency from acquisition to product generation is reduced to minutes or hours, enabling faster situational awareness for emergency response.

\subsection{Improved Detection of Fragmented Fires}

The spatial resolution of 3~m and \ac{MMU} of 3~ha significantly enhance the ability to detect small or discontinuous burn scars in heterogeneous landscapes. For national authorities such as ISPRA, SNPA, and the Civil Protection Department, this capability is essential for legal enforcement and ecological damage assessment in wildland-urban interfaces~\cite{Taramelli2021_MirrorCopernicus}.

\subsection{Implications for Onboard Model Design}

Deploying \ac{AI} models onboard radiation-constrained, low-power satellite platforms such as the iX5 Myriad-X requires careful optimisation of model size, execution time and robustness. Addressing these constraints through hardware-aware model and pipeline co-design enables reliable, real-time inference in orbit, ensuring that onboard processing delivers consistent service-level benefits within the operational \ac{IRIDE} \ac{SVC}.

\section{Designing Onboard AI Models for Operational EO Services}

Translating operational service requirements into deployable onboard AI requires a system-aware approach to AI model design. Unlike ground-based processing pipelines, onboard inference must operate under strict constraints on memory footprint, computational throughput, power consumption, and execution latency. As a result, the feasibility of onboard processing for operational EO services is determined not only by algorithmic accuracy, but by the joint optimisation of models and target hardware.

\subsection{Deployment Constraints and Requirements}

The onboard \ac{VPU} adopted within the \ac{HEO} constellation, the Intel\textsuperscript{\textregistered} Myriad X \ac{VPU}, is representative of a class of low-power accelerators designed for embedded and spaceborne inference. While capable of high-throughput execution, such platforms impose hard limits on model size and require support for reduced-precision arithmetic (e.g.\ INT8 or FP16). Inference latency must remain compatible with real-time or near-real-time product generation to ensure that onboard processing delivers tangible service-level benefits.

Meeting these constraints typically requires a combination of optimisation strategies, including quantisation, pruning and knowledge distillation, and approaches like \ac{NAS}, enabling automated co-design of model structure and deployment characteristics. 

The \ac{HEO} onboard AI stack leverages inference toolchains such as OpenVINO, supporting model conversion, layer fusion, and runtime optimisation for deployment on the Myriad X accelerator. These tools provide a practical bridge between algorithm development and operational onboard execution.

\subsection{From Feasibility to Service Integration}

Recent work \cite{DelPrete2025} demonstrated that hardware-aware NAS can generate lightweight segmentation models satisfying the latency and resource constraints of Myriad and Orin-class processors, achieving substantial speedups with respect to manually designed baselines. While that study focused on methodological validation and benchmarking across hardware platforms, it establishes the technical feasibility required for deploying high-resolution segmentation models in orbit.

Within the scope of the present paper, this class of hardware-aware design approaches is considered an enabling capability rather than a delivered service component. The focus remains on assessing the architectural and operational conditions under which onboard processing provides added value at the service level, independent of any specific model instantiation.

\section{Conclusion and Outlook}

This paper assessed onboard processing as a service-level enabler within operational \ac{EO} architectures, using the \ac{IRIDE} \ac{HEO} service segment and burnt-area mapping as a case study. Relative to ground-only pipelines, hybrid onboard-ground chains can reduce latency and downlink load while improving continuity under emergency acquisition peaks. The analysis positions \ac{IRIDE} \ac{HEO} as complementary to Copernicus services, targeting higher spatial detail and smaller event detectability at national scale. Ongoing work focuses on integrating a deployable onboard burnt-area model into the \ac{IRIDE} service value chain, leveraging hardware-aware design approaches such as NAS demonstrated in~\cite{DelPrete2025}.

\small
\bibliographystyle{IEEEtranN}
\bibliography{references}

\end{document}